\documentclass{ecai}
\usepackage{times}
\usepackage{epsfig}
\usepackage{graphicx}
\usepackage{amsmath}
\usepackage{amssymb}
\usepackage{times}
\usepackage{soul}
\usepackage[hidelinks]{hyperref}
\usepackage[utf8]{inputenc}
\usepackage[small]{caption}
\usepackage{graphicx}
\usepackage{amsmath}
\usepackage{booktabs}
\usepackage{multirow}
\usepackage{amsfonts}
\usepackage{multirow}
\usepackage{color}
\usepackage[table]{xcolor}
\definecolor{lightgray}{gray}{0.9}
\definecolor{grey1}{gray}{0.8}
\title{SSHFD: Single Shot Human Fall Detection with Occluded Joints Resilience}
\author{Umar Asif \institute{IBM Research Australia, email: umarasif@au1.ibm.com} \and Stefan Von Cavallar \institute{IBM Research Australia, email: svcavallar@au1.ibm.com} \and Jianbin Tang \institute{IBM Research Australia, email: jbtang@au1.ibm.com} \and Stefan Harrer\institute{IBM Research Australia, email: sharrer@au1.ibm.com}}
\begin{document}
\maketitle
\begin{abstract}
Falling can have fatal consequences for elderly people especially if the fallen person is unable to call for help due to loss of consciousness or any injury.
Automatic fall detection systems can assist through prompt fall alarms and by minimizing the fear of falling when living independently at home. 
Existing vision-based fall detection systems lack generalization to unseen environments due to challenges such as variations in physical appearances, different camera viewpoints, occlusions, and background clutter. 
In this paper, we explore ways to overcome the above challenges and present Single Shot Human Fall Detector (SSHFD), a deep learning based framework for automatic fall detection from a single image.
This is achieved through two key innovations. First, we present a human pose based fall representation which is invariant to appearance characteristics. Second, we present neural network models for 3d pose estimation and fall recognition which are resilient to missing joints due to occluded body parts. 
Experiments on public fall datasets show that our framework successfully transfers knowledge of 3d pose estimation and fall recognition learnt purely from synthetic data to unseen real-world data, showcasing its generalization capability for accurate fall detection in real-world scenarios.
\end{abstract}
\section{Introduction}
Falling on the ground is considered to be one of the most critical dangers for elderly people living alone at home which can cause serious injuries and restricts normal activities because of the fear of falling again \cite{fleming2008inability}. Automated fall detection systems can produce prompt alerts in hazardous situations.
They also allow automatic collection and reporting of fall incidents which can be used to analyse the causes of falls, thus improving the quality of life for people with mobility constraints and limited supervision.
Vision-based systems provide a low cost solution to fall detection. They do not cause sensory side effects on the human health and do not affect the normal routines of elderly people as observed in systems using wearable devices \cite{zhao2012exploration}.
In a typical fall detection approach, human regions are detected from the visual data and used to learn features to distinguish fall from other activities.
Existing methods such as \cite{mirmahboub2013automatic} learn fall representations using physical appearance based features extracted from video data. 
However, appearance based features suffer from poor generalization in real-world environments due to large variations in appearance characteristics, different camera viewpoints, and background clutter. 
Furthermore, due to the unavailability of large-scale public fall datasets, most of the existing fall detectors are trained and evaluated using simulated environments or using restricted datasets (which cannot be shared publicly due to privacy concerns). Therefore, these methods do not exhibit generalization capabilities for fall detection in unseen real-world environments.
In this paper, we explore ways to overcome the above challenges and present a deep learning framework termed ``Single Shot Human Fall Detector (SSHFD)" for accurate fall detection in unseen real-world environments.
The main contributions of this paper are as follows:
\begin{enumerate}
\item
We present a human pose based fall representation which is invariant to appearance characteristics, backgrounds, lighting conditions, and spatial locations of people in the scene. Experiments show that neural network models trained on our 2d-pose and 3d-pose based fall representations successfully generalize to unseen real-world environments for fall recognition.
\item
We present neural network models for 3d pose estimation and fall recognition which are robust to partial occlusions. Experiments show that our models successfully recover joints information from occluded body parts, and accurately recognize fall poses from incomplete input data. 
\item
We evaluate our framework on real-world public fall datasets, where we show that our framework when trained using only synthetic data, shows excellent generalization capabilities of fall recognition on unseen real-world data.
\end{enumerate}
\section{Related Work}\label{related_work}
\begin{figure*}[t!]
	\begin{center}
		\includegraphics[trim=0.0cm 0.0cm 0.0cm 0.0cm,clip,width=1.0\linewidth,keepaspectratio]{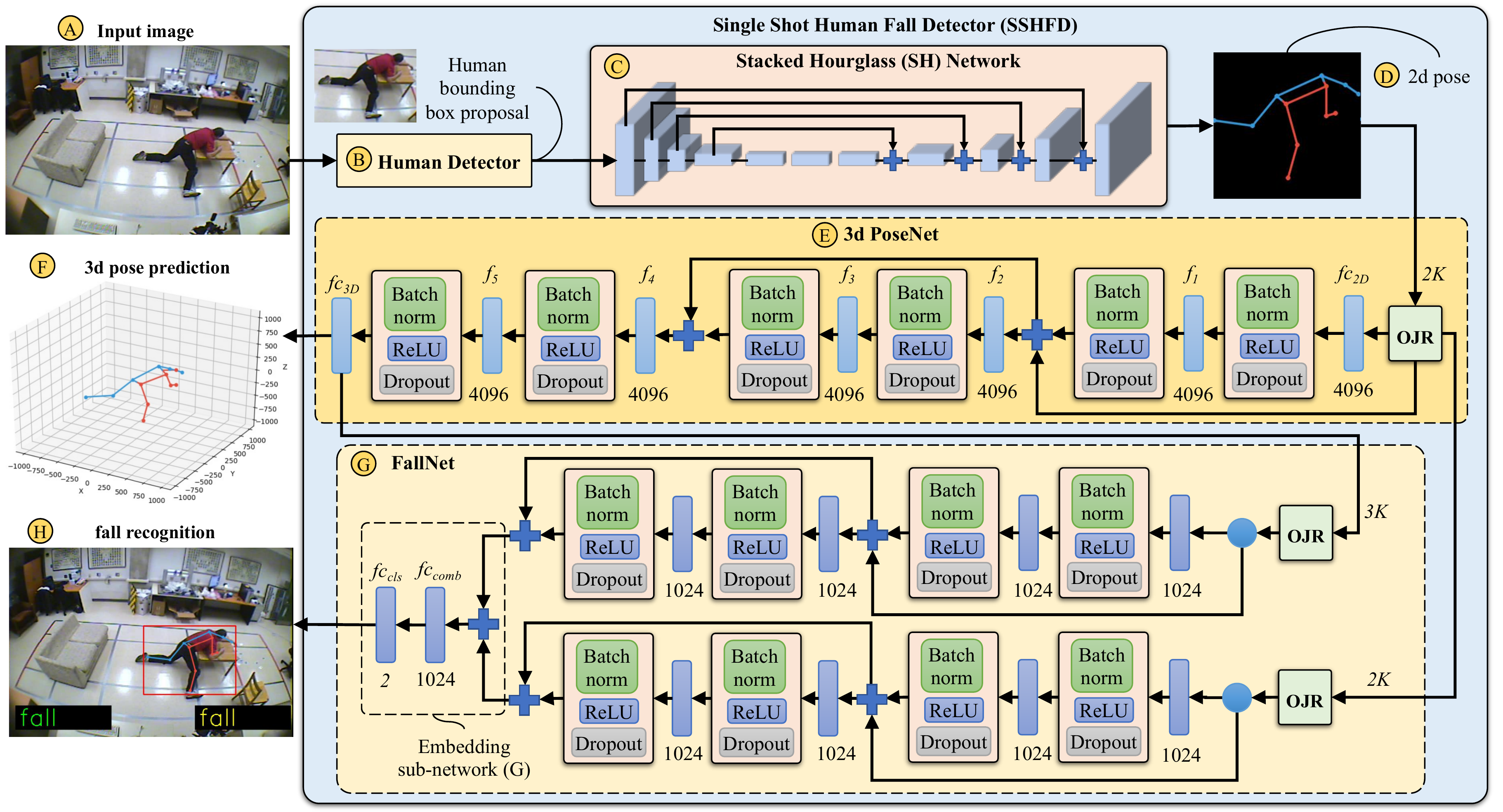}
		\vspace{-12pt}
		\caption{Overview of our Single Shot Human Fall Detector (SSHFD). Given a single RGB image of the scene (A), SSHFD  generates human proposals (B) which are fed into a Stacked Hourglass network (C) for 2d pose prediction. Next, the predicted 2d pose (D) is fed into a neural network (E) for 3d pose prediction (F). Finally, the 2d pose and the 3d pose information are fed into a neural network (G) for fall recognition (H). Our models integrate Occluded Joints Resilience (OJR) modules which make the models robust to missing information in the pose data.}
		\vspace{-8pt}
		\label{fig_framework}
	\end{center}
\end{figure*}
Existing vision-based fall detection approaches detect human regions in the scene and use visual information from the detected regions to learn features for fall recognition. For instance, the method of \cite{miaou2006customized} generated human bounding boxes through background-foreground subtraction and compared the visual content of the boxes in consecutive frames of the videos of the MultiCam fall dataset \cite{auvinet2010multiple} to detect fall events. 
The method of \cite{qian2008home} compared multiple bounding boxes to distinguish between different events (e.g., standing, sitting, and fall). 
The work of \cite{juang2007human} used a fuzzy neural network classifier for fall detection.
The methods of \cite{mirmahboub2013automatic} and \cite{huang2004extreme} used motion segmentation to detect human regions in the scene and combined visual appearance and shape information from the detected regions to learn features for fall recognition. 
However, errors in background-foreground subtraction or motion segmentation (e.g., due to small or no change in the visual content between subsequent image frames) degrade the accuracy of these methods.
To overcome this challenge, the method of \cite{hung2013detecting} used cues from multiple cameras and produced fall decisions through voting among different viewpoints. However, this approach requires accurate synchronization between the individual cameras.
Other methods such as  \cite{gasparrini2014depth,mastorakis2014fall} used Kinect depth maps to learn 3d features for fall recognition. 
However, these methods are restricted in real-world deployment due to hardware limitations (e.g. limited depth sensing range).
Compared to existing methods, our work differs in several ways.
\textbf{First}, our framework learns pose based fall representations which are invariant to appearance characteristics. This enables our framework to successfully transfer fall recognition knowledge learned from pure synthetic data to real-world data with unknown backgrounds and different human actors.
\textbf{Second}, our framework integrates a 3d pose estimator which predicts 3d pose information from 2d pose. The combined 2d and 3d pose knowledge enables our framework to successfully handle ambiguities in the 2d pose (under different camera viewpoints), without requiring multiple camera setups or depth sensor technologies. 
\textbf{Finally}, our neural network models for 3d pose estimation and fall recognition are resilient to missing information in the pose data. This enables our framework to accurately discriminate between fall and no-fall cases from human poses under occlusions. 
\section{The Proposed Framework (SSHFD)}\label{proposed_framework}
Fig. \ref{fig_framework} shows the overall architecture of our framework which has three main modules. \textbf{i)} 2d pose estimation, which takes an RGB image of the scene as input and produces body joints locations in 2d image space, \textbf{ii)} 3d pose estimation, which takes 2d pose as input and predicts joints locations in 3d Cartesian space, and \textbf{ii)} Fall recognition, which combines 2d pose and 3d pose data and predicts probabilities with respect to the target classes. In the following, we describe in detail the individual components of our framework.
\subsection{The Proposed Fall Representation}
Our fall representation is based on joints locations in 2d image space and 3d Cartesian space. We normalize the 2d pose by transforming the joints estimates (predicted in the scene image) to a fixed reference image of $224\times 224$ dimensions as shown in Fig. \ref{fig_framework}-D. The normalized 2d pose is then used to predict joints locations in a Cartesian space of size $1000\times 1000\times 1000mm^3$ as shown in Fig. \ref{fig_framework}-F. The 3d predictions are normalized with respect to the hip joint.
\subsection{The Proposed 2d Pose Estimation (Fig. \ref{fig_framework})}\label{pose_estimation}
Our 2d pose estimator is composed of two main modules: \textbf{i)} a human detector \cite{Detectron2018},  which produces human bounding box proposals from the input image, and \textbf{ii)} a Stacked Hourglass (SH) network \cite{newell2016stacked}, which predicts body joints 2d locations and their corresponding confidence scores. The SH network is trained using ground truth labelled in terms of $W\times H\times K-$dimensional heatmaps ($\mathcal{H}$), where $W$ and $H$ represent the width and height of the heatmap and $K$ represents the number of joints. We used $K=17$ joint types as per the format used in \cite{Detectron2018}. The heatmap ($\mathcal{H}_{k}$) for a joint $k\in\{1,...,K\}$ is generated by centering a Gaussian kernel around the joint's pixel position $(x_k,y_k)$. It is given by:
\begin{equation}\label{eq_heatmap}
\mathcal{H}_{k}(x,y)=\frac{1}{2\pi \sigma^{2}}\exp(\frac{-[(x-x_{k})^2+(y-y_{k})^2]}{2\sigma^{2}}),
\end{equation}
where $\sigma$ is a hyper-parameter for spatial variance. We set $\sigma=4$ in our experiments. 
The training objective function of the SH network is defined by:
\begin{equation}\label{eq_loss_pose_2d}
\mathcal{L}_{2d}=\frac{1}{K}\sum_{k=1}^{K}||\mathcal{H}_{k}-\hat{\mathcal{H}}_{k}||_{2}^{2},
\end{equation}
where $\hat{\mathcal{H}}_{k}$ represents the predicted confidence map for the $k$th joint.
\subsection{The Proposed 3D Pose Estimation (Fig. \ref{fig_framework}-E)}
Here, the goal is to estimate $K$ body joints in 3d Cartesian space $\boldsymbol{Q}\in \mathbb{R}^{3K}$ given a 2d input $\boldsymbol{P}\in \mathbb{R}^{2K}$. For this, we learn an objective function $\mathcal{F}^{*}:\mathbb{R}^{2K}\rightarrow \mathbb{R}^{3K}$ which minimizes the prediction error over a dataset with $N$ poses:
\begin{equation}\label{loss_ggpn}
\mathcal{F}^{*}=\min_{f}\frac{1}{N}\sum_{i=1}^{N}\mathcal{L}_{3d}(f(\boldsymbol{p}_{i})-\boldsymbol{q}_{i}),
\end{equation}
where $\mathcal{L}_{3d}$ represents an MSE loss.
Fig. \ref{fig_framework}-E shows the structure of our 3d pose estimation model \textit{``3d PoseNet"} based on the architecture of \cite{martinez2017simple}. It starts with a linear layer $fc_{2D}$ which transforms the $2K-$dimensional pose to $1024$ dimensional features. Next, there are five linear layers $f_{1}-f{5}$, each with $4096$ dimensions followed by Batch normalization, a Rectified Linear Unit and a dropout module. The final layer $fc_{3D}$ produces $3K$ dimensional ouptut. There are two residual connections defined in the network which combine information from lower layers to higher layers and improve model generalization performance. 
\subsection{The Proposed Fall Recognition (Fig. \ref{fig_framework}-G)}\label{fallnet}
We present a neural network (\textit{FallNet}) which consists of two sub-networks: a modality-specific network $F_{\phi}, \phi\in\{\boldsymbol{P},\boldsymbol{Q}\}$, and an embedding network $\boldsymbol{G}$ as shown in Fig. \ref{fig_framework}-G.
The sub-network $F_{\phi}$ has a structure similar to \cite{martinez2017simple} but with fewer linear layers. It produces $1024-$dimensional features each from the two input modalities ($\boldsymbol{P}$ and $\boldsymbol{Q}$). The output features are summed and fed into the embedding sub-network $\boldsymbol{G}$ which uses two linear layers and learns probabilistic distributions with respect to the target classes.
Let $\rho_{i}$ denote the outputs of the last layer ($fc_{cls}$) for the $i^{th}$ input sample. The training objective function is defined over ${N}$ poses as:
\begin{equation}\label{loss_ggpn}
\mathcal{L}_{fall}=\sum_{i\in {N}}\mathcal{L}_{cls}(\rho_{i},\rho_{i}^{*}),
\end{equation}
where $\rho^{*}_{i}$ represent the ground-truth labels. 
The term $\mathcal{L}_{cls}$ is a Cross Entropy Loss, given by:
\begin{equation}\label{loss_cls}
\mathcal{L}_{cls}(\boldsymbol{\textup{x}},C)=-\sum_{C=1}^{N_{C}}\mathcal{Y}_{\textup{x},C}\log(p_{\textup{x},C}),
\end{equation}
where $\mathcal{Y}$ is a binary indicator if class label $C$ is the correct classification for observation $\textup{x}$, and $p$ is the predicted probability of observation $\textup{x}$ of class $C$.
\subsection{The Proposed Occluded Joints Resilience (OJR)}\label{section_ojr}
\begin{figure}[t!]
	\begin{center}
		\includegraphics[trim=0.0cm 0.2cm 0.0cm 0.0cm,clip,width=1.0\linewidth,keepaspectratio]{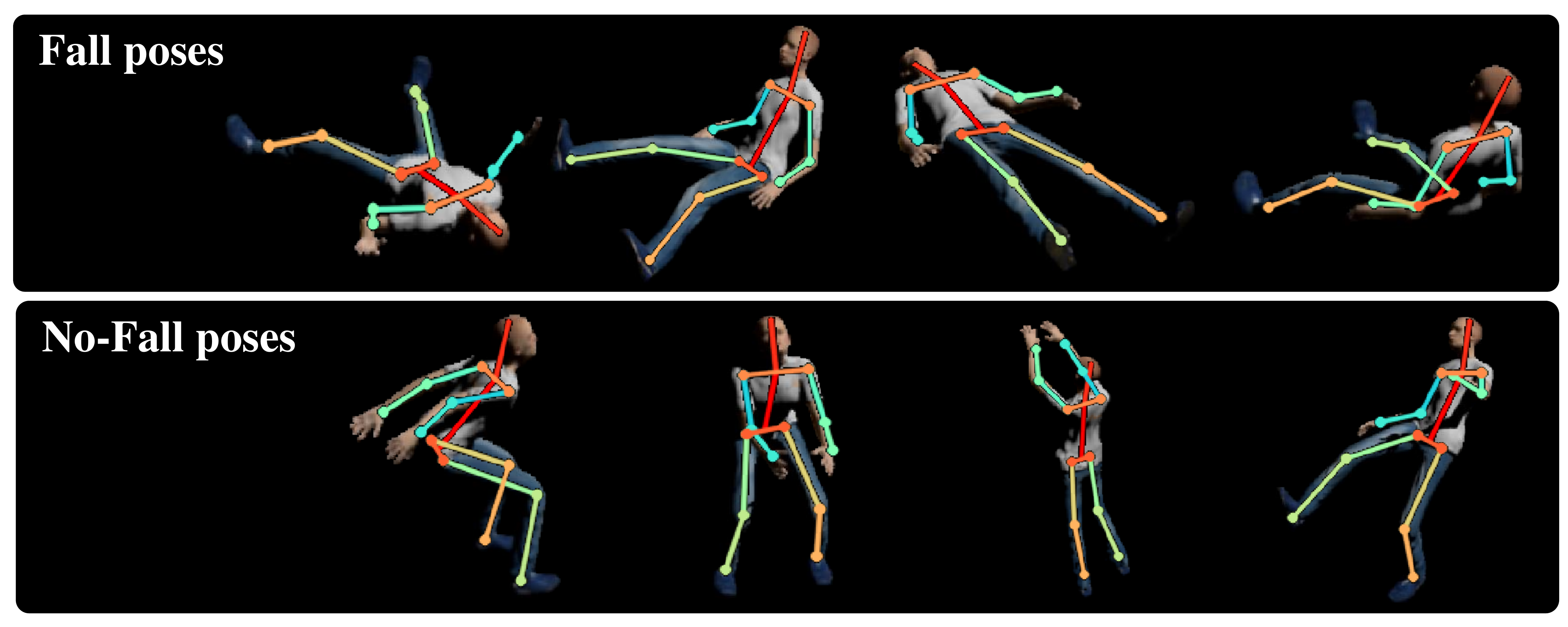}
		\vspace{-5pt}
		\caption{Sample frames from our Synthetic Human Fall dataset showing different poses.}
		\vspace{-5pt}
		\label{fig_dataset}
	\end{center}
\end{figure}
\definecolor{grey1}{gray}{0.8}
\begin{table*}[t!]
	\caption{Fall recognition results of the proposed SSHFD in terms of its different variants termed Human Fall Detection Models (HDF) on the MultiCam fall dataset and the Le2i fall dataset. The models for 3d pose estimation and fall recognition were trained only on the synthetic data and evaluated on real-world test datasets.}
	\vspace{-0pt}	
	\centering
	\setlength\tabcolsep{5.0pt}\centering
	\begin{tabular}{@{}lccccccccc@{}}
		\toprule
		\multirow{2}{*}{Human Fall Detection (HFD) Models}&\multicolumn{3}{c}{MultiCam fall dataset}&\multicolumn{3}{c}{Le2i fall detection database}\\
		&\multirow{1}{*}{F1Score}&\multirow{1}{*}{Precision}&\multirow{1}{*}{Recall}& {F1Score}&\multirow{1}{*}{Precision}&\multirow{1}{*}{Recall}\\	
		\midrule
		\multirow{1}{*}{SSHFD-A: SH + FallNet2d3d}		
		&	{0.8453}		&{0.8487}	&{0.8431} 	
		&{0.8991}		&{0.9008}	&{0.8992}
		\\
		\multirow{1}{*}{SSHFD-B: SH + FallNet2d}			
		&0.8388 & 0.8437 & 0.8358
		&0.8885 & 0.8907 & 0.8887
		\\
		\multirow{1}{*}{SSHFD-C: SH + ResNet (RGB)}		 
		&{0.8638}& {0.8628} &{0.8658}
		&0.6595 & 0.7985 & 0.6912\\
		\bottomrule
	\end{tabular}
	\vspace{-0pt}
	\label{table_detection1}
\end{table*}
\begin{figure*}[t!]
	\begin{center}
		\includegraphics[trim=0.2cm 1.0cm 0.2cm 0.4cm,clip,width=1.0\linewidth,keepaspectratio]{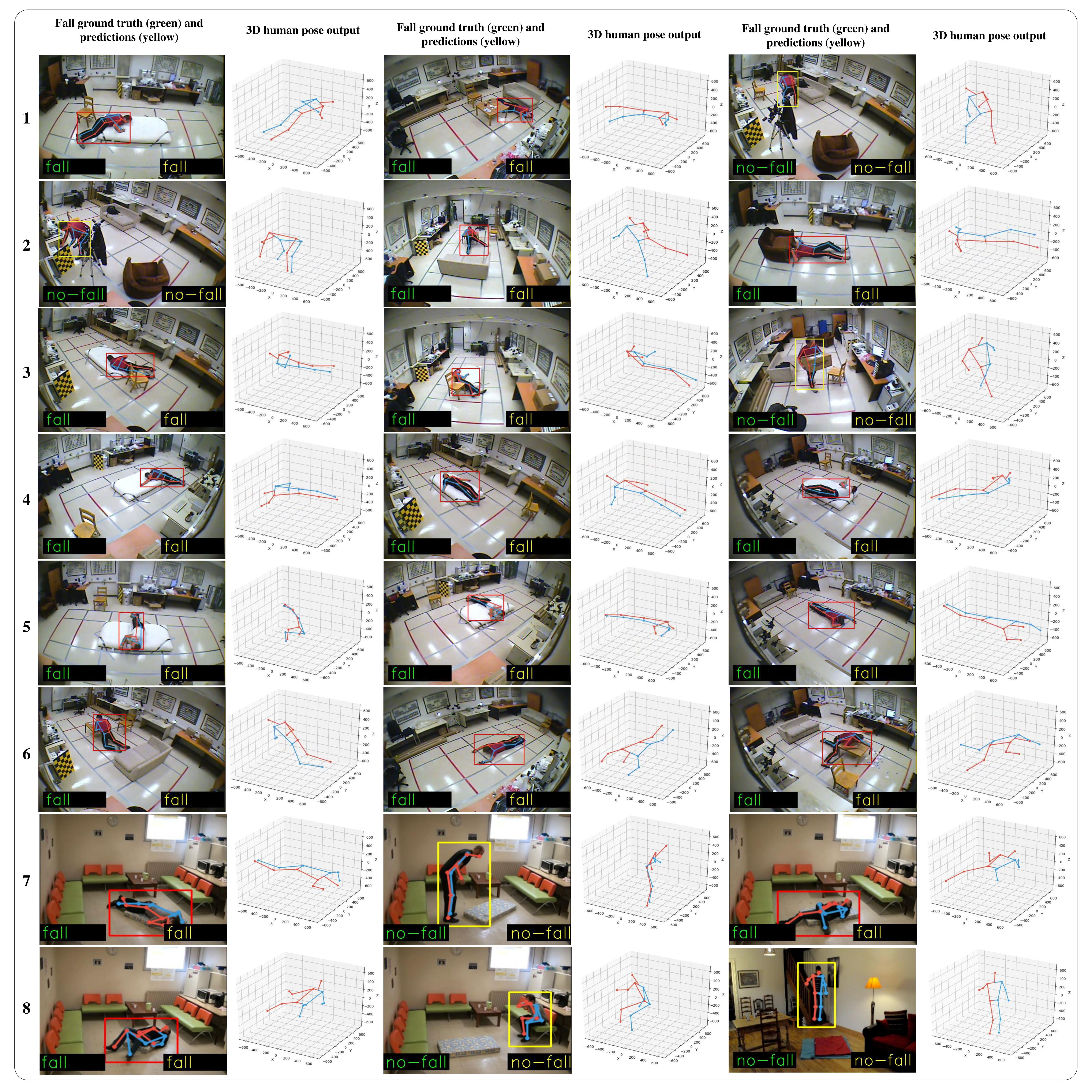}
		\vspace{-12pt}
		\caption{Qualitative results of our framework on the MultiCam fall dataset (rows 1-6) and the Le2i fall dataset (rows 7-8). Ground truth labels and model predictions are shown by text in green and yellow, respectively. Fall and no-fall cases are represented by bounding boxes in red and yellow, respectively.}
		\label{fig_results}
		\vspace{-5pt}
	\end{center}
\end{figure*}
\begin{figure*}[t!]
	\begin{center}
		\includegraphics[trim=0.0cm 0.0cm 0.0cm 0.4cm,clip,width=1.0\linewidth,keepaspectratio]{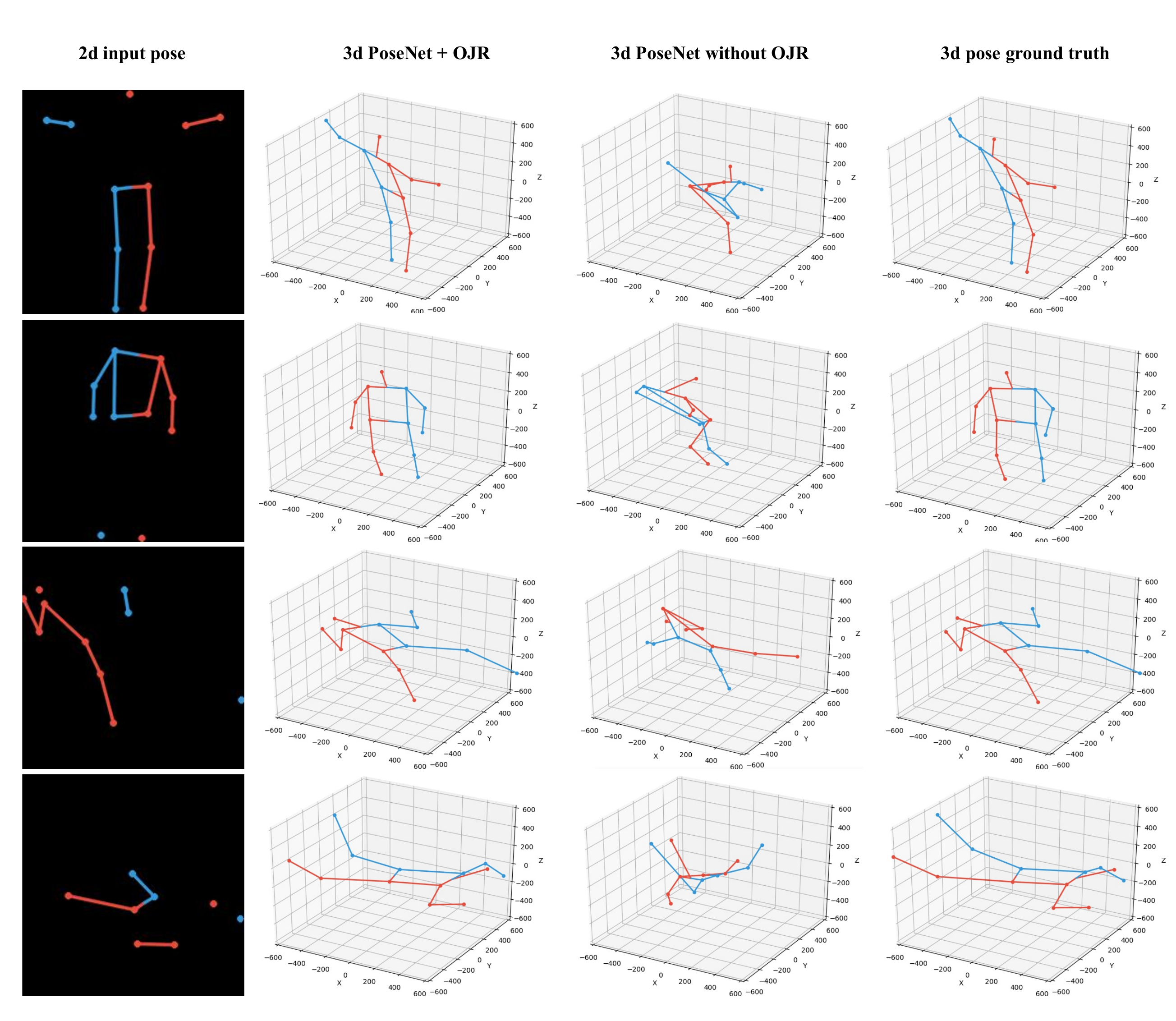}
		\vspace{-15pt}
		\caption{Qualitative comparison of the predictions of our 3d PosNet with and without the proposed OJR using inputs with missing joints on our synthetic dataset. Our OJR-based 3d PoseNet enables the model to successfully recover missing joints in the input pose data.}
		\label{fig_human}
		\vspace{-10pt}
	\end{center}
\end{figure*}
\begin{figure*}[t!]
	\begin{center}
		\includegraphics[trim=0.1cm 0.0cm 0.1cm 0.0cm,clip,width=0.97\linewidth,keepaspectratio]{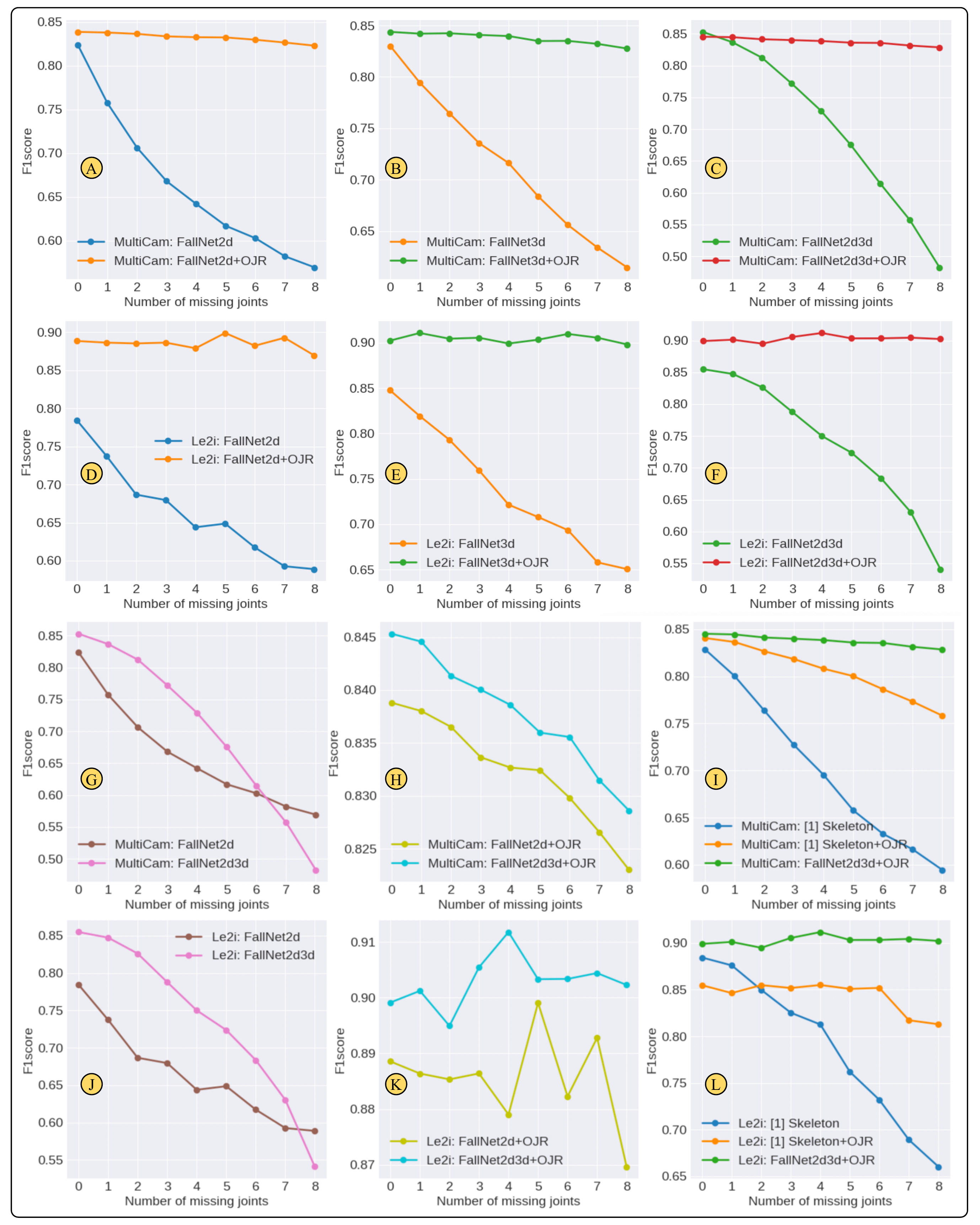}
		\vspace{-5pt}
		\caption{F1scores of our SSHFD on the MultiCam fall dataset and the Le2i fall detection database under different noise levels. The subplots A-F show that the proposed OJR-based models produce considerable higher f1scores for fall recognition under missing joints information compared to the models which were trained without the proposed OJR method. The subplots G, H, J, and K, show comparison between our 2d-pose based model ``FallNet2d" and ``FallNet2d3d" which uses both 2d and 3d pose for fall recognition. The subplots I and L show a comparison between f1scores of our method and the visual skeleton representation based method of \cite{asifprivacy} under different noise levels.}
		\label{fig_noise_levels}
		\vspace{-15pt}
	\end{center}
\end{figure*}
\begin{figure}[t!]
	\begin{center}
		\includegraphics[trim=0.4cm 0.4cm 0.4cm 0.0cm,clip,width=1.0\linewidth,keepaspectratio]{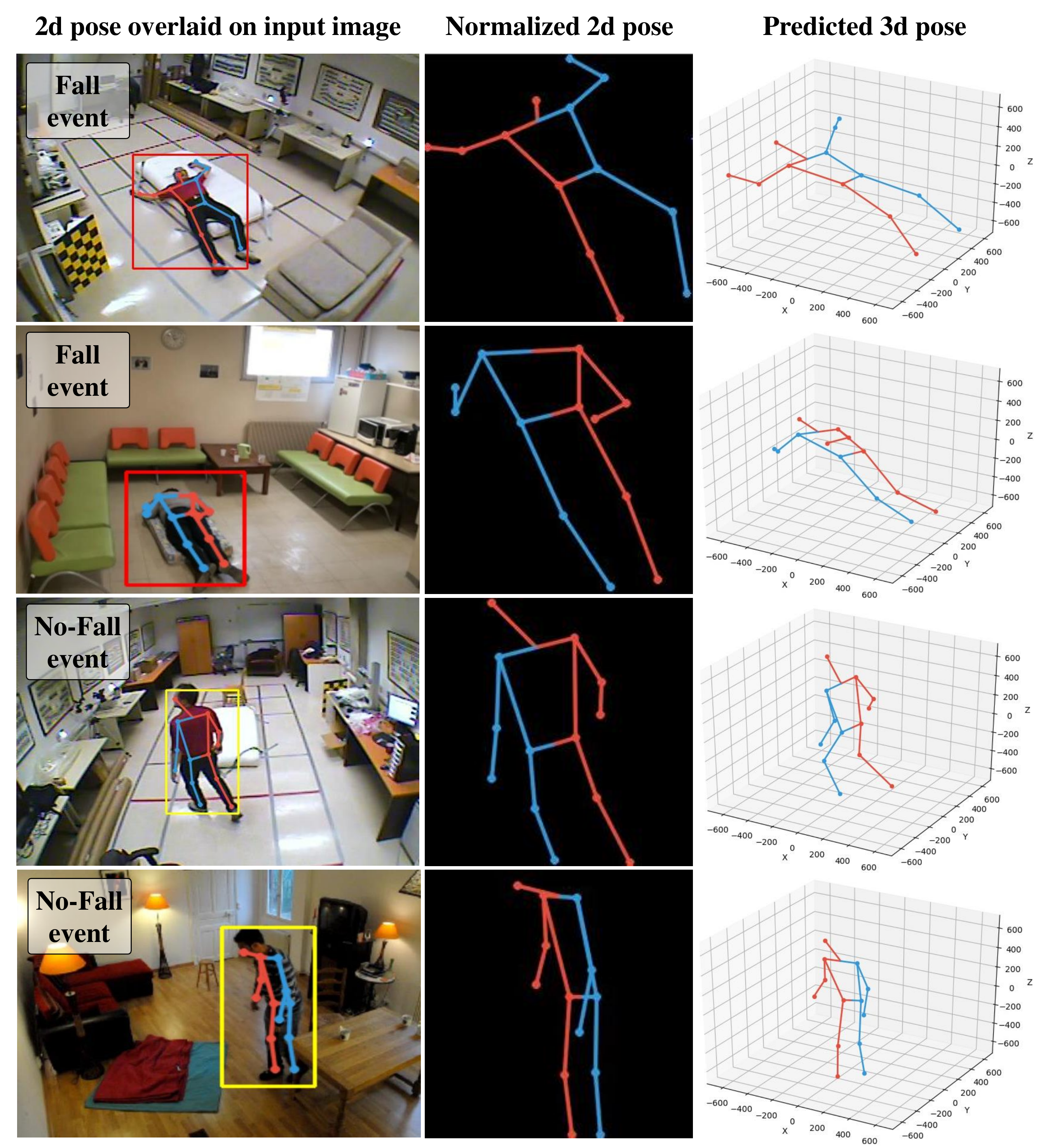}
		\vspace{-10pt}
		\caption{Variations in camera viewpoints cause ambiguities in 2d pose based fall representations (fall and no-fall 2d poses resemble each other as shown in the middle column). In contrast, our 3d PoseNet predictions (as shown in the right column) are more discriminative and reduce inter-class similarities for fall recognition. Fall and no-fall cases are represented by bounding boxes in red and yellow, respectively.}
		\vspace{-5pt}
		\label{fig_ambiguities}
	\end{center}
\end{figure}
Pose estimators trained on RGB images inevitably make errors in joint predictions due to factors such as: image imperfections, occlusions, background clutter, and incorrect ground truth annotations. Since, our 3d PoseNet and FallNet models rely on the output of the SH network, errors in 2d pose predictions affect the quality of 3d pose estimation and fall recognition.
To overcome this challenge, we present a method termed ``Occluded Joints Resilience (OJR)" which increases the robustness of our models to incomplete information in the pose data.
To achieve this, the OJR method creates an occlusion pattern $\mathcal{M}_i$ and uses it to transform the original pose data into occluded pose data. The occlusion pattern $\mathcal{M}_i$ is defined as:
\begin{equation}\label{eq_occ}
\mathcal{M}_i=[v_1J_1,...,v_kJ_k], v\in \{0,1\},
\end{equation}
where $J_{i}=(x_i,y_i)$ represents a body joint and $v$ is a binary variable, indicating the visibility of the $k$th joint. 
During training, the OJR method generates a rich library of unique occlusion patterns $\{\boldsymbol{\mathcal{M}}\}$ which vary across training samples, thereby increasing the network's adaptivity to various occluded situations.
\section{Experiments}\label{experiments}
\subsection{Training and Implementation Details}\label{implementation}
We trainined the SH network for 2d pose estimation using the MS  COCO Keypoints dataset \cite{lin2014microsoft}, which contains 64K images and 150K instances with 2d pose ground truth. 
To train our models for 3d pose estimation and fall recognition, we used the synthetic human fall dataset of \cite{asifprivacy}, which provides 767K samples of human poses with 2d and 3d pose annotations categorized into fall and no-fall body poses. 
Fig. \ref{fig_dataset} shows some samples from the synthetic dataset.
For training the 3d PoseNet and FallNet models, we initialized the weights of the fully connected layers with zero-mean Gaussian distributions (standard deviations were set to 0.01 and biases were set to 0), and trained each network for 300 epochs. The starting learning rate was set to 0.01 and divided by 10 at 50\% and 75\% of the total number of epochs. The parameter decay was set to 0.0005 on the weights and biases. The probability of dropout was set to 0.5.
Our implementation is based on the framework of Torch library \cite{paszke2017automatic}. Training was performed using ADAM optimizer and four Nvidia Tesla K80 GPUs. 
\subsection{Test Datasets}
To evaluate the generalization capability of our SSHFD for fall detection in unseen real-world environments, we trained our models using only synthetic data and tested the models on the public MultiCam fall dataset \cite{auvinet2010multiple} and the Le2i fall dataset \cite{charfi2013optimized}. The MultiCam dataset consists of 24 different scenarios where each scenario is comprised of a video sequence of a person performing a number of activities (such as falling on a mattress, walking, carrying objects). Each scenario is recorded using 8 cameras from 8 different locations. The dataset is challenging for single-shot single camera fall detection because, different camera viewpoints produce occlusions and significant variations in the spatial locations, scale, and orientations of the falls \cite{asifprivacy}. The Le2i dataset contains 221 videos of different actors performing fall actions and various other normal activities in different environmets. The dataset is challenging due to variable lighting conditions and occlusions \cite{asifprivacy}.
To quantify the recognition performance of our SSHFD, we extracted image frames from the target videos at 25 fps resolution and generated 2d poses using the SH network. Next, we computed weighted F1 scores, precision (PRE) and recall (RE) scores per image frame with atleast one pose detected and averaged the scores over all image frames of the targets datasets.  
We used the weighted measures as they are not biased by imbalanced class distributions which make them suitable for the target datasets where the number of fall samples are considerably small compared to the number of non-fall samples. 
\section{Results}\label{results}
Table \ref{table_detection1} shows fall recognition results on the test datasets, for different variants of our framework termed ``Human Fall Detection Models". 
The variants ``A" and ``B" use neural networks with linear structures which were trained on pose data as shown in Fig. \ref{fig_framework} and described in Sec. \ref{fallnet}. 
The variant ``C" shown in Table \ref{table_detection1} uses a ResNet18 \cite{he2016deep} based CNN architecture which was trained on RGB appearance information of synthetic human proposals.
The results reported in Table \ref{table_detection1} show that although the RGB-based fall detector produced higher f1scores compared to the pose-based fall detectors on the MultiCam dataset, it produced the lowest f1scores on the Le2i dataset.
This is because, the RGB-based fall detector trained on color information of synthetic human proposals failed to generalize to the scenes of Le2i dataset with high variations in the appearance characteristics and different backgrounds. 
Compared to the RGB-based detector, our pose-based fall detector (SSHFD-A) produced competitive f1scores on the MultiCam dataset and superior f1scores on the Le2i dataset as shown in Table \ref{table_detection1}.
Fig. \ref{fig_results} shows qualitative results of our pose-based fall detector on sample images from the test datasets. The results show that our fall detection framework is robust to partial occlusions, and variations in the spatial locations, scale, and orientations of fall poses in real-world scenes. These improvements are attributed to our pose-based fall representation which is invariant to appearance characteristics and makes our framework robust to different human actors and background clutter in real-world scenes. 
These results demonstrate the generalization capability of our framework in successfully transferring fall recognition knowledge learnt purely from synthetic data to unseen real-world data.
Table \ref{table_detection1} also shows that our FallNet2d3d model using combined 2d- and 3d-pose information performed better than the FallNet2d model which used only 2d pose information. This is attributed to the proposed FallNet architecture which uses low-level modality-specific layers to learn discriminative information from the individual pose modalities, and uses high-level fusion layers to learn the complimentary information in the multi-modal input pose data, thereby producing features which are robust to pose ambiguities in the 2d image space under different camera viewpoints as shown in Fig. \ref{fig_ambiguities}.
\subsection{Robustness to Missing Joints}
\begin{figure}[t!]
	\begin{center}
		\includegraphics[trim=0.4cm 0.4cm 0.4cm 0.0cm,clip,width=1.0\linewidth,keepaspectratio]{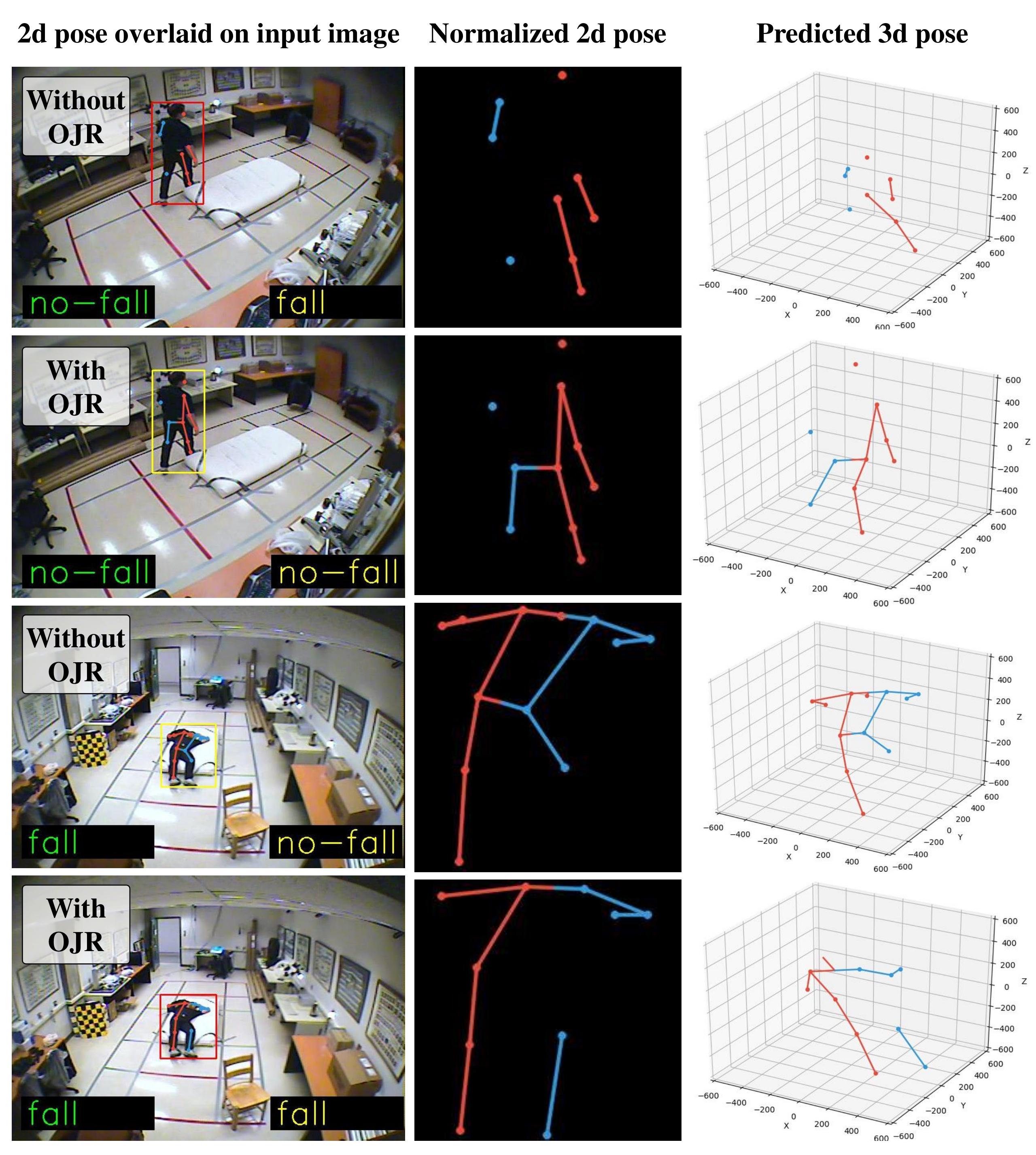}
		\vspace{-10pt}
		\caption{Our OJR-enabled FallNet model produces correct fall predictions in the presence of missing data in 2d pose and 3d pose compared to the model which was trained without the OJR method. Ground truth labels and model predictions are shown by text in green and yellow, respectively. Fall and no-fall cases are represented by bounding boxes in red and yellow, respectively.}
		\vspace{-15pt}
		\label{fig_missing_3d}
	\end{center}
\end{figure}
\begin{figure}[t!]
	\begin{center}
		\includegraphics[trim=0.4cm 0.4cm 0.4cm 0.0cm,clip,width=1.0\linewidth,keepaspectratio]{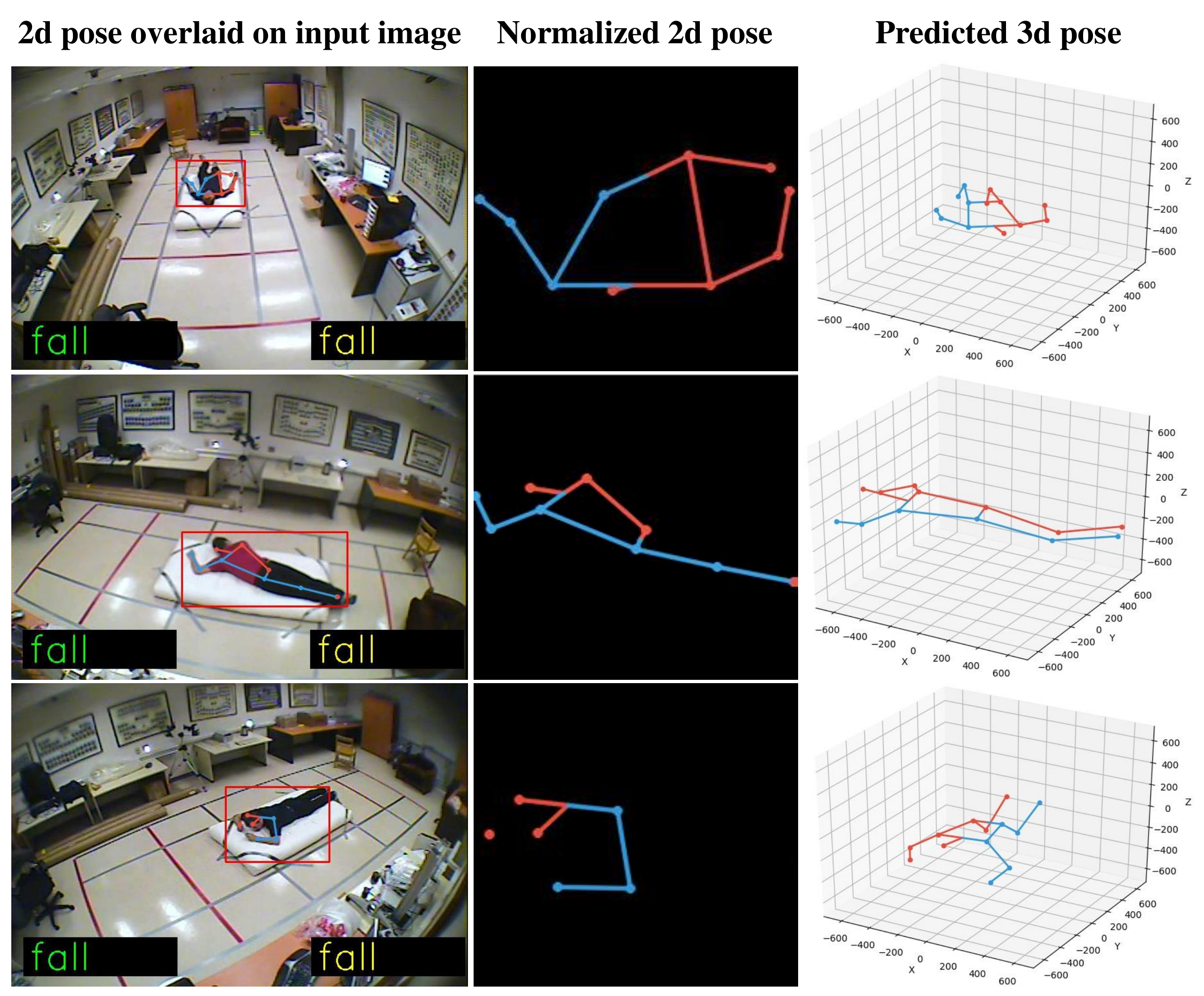}
		\vspace{-10pt}
		\caption{Our 3d PoseNet using the proposed OJR method successfully recovers missing data in the input 2d pose and enables the framework to produce correct fall predictions. Ground truth labels and model predictions are shown by text in green and yellow, respectively.}
		\vspace{-15pt}
		\label{fig_missing_joints}
	\end{center}
\end{figure}
\subsubsection{Fall Recognition}
Fig. \ref{fig_noise_levels} shows comparison of f1scores produced by our models with and without the proposed OJR  on the MultiCam dataset and the Le2i fall dataset under different noise levels. The results show that our OJR-based models produced significantly higher f1scores for all the noise levels compared to the models which were trained without using OJR. For instance, using input pose data with 8 missing joints, our OJR-based models improved f1scores by upto 35\% and 40\% compared to the models without using OJR on the MultiCam and Le2i datasets, respectively (see Fig. \ref{fig_noise_levels}-C and Fig. \ref{fig_noise_levels}-F). 
We also conducted experiments to compare the performance of our FallNet with the method of \cite{asifprivacy} which uses visual representations of 2d skeletons and segmentation information for fall recognition. Fig. \ref{fig_noise_levels}-I and Fig. \ref{fig_noise_levels}-L show the results of these experiments. The results show that our 2d- and 3d-pose based fall representation produces superior fall recognition performance especially under missing joints data compared to the skeleton-based visual representation of \cite{asifprivacy}. Fig. \ref{fig_noise_levels}-I and Fig. \ref{fig_noise_levels}-L also show that the proposed OJR improved the performance of the method of \cite{asifprivacy} under different noise levels, demonstrating the significance of the proposed OJR for improving the robustness of models under scenarios with occluded joints.
Fig. \ref{fig_missing_3d} shows qualitative results of our models on the MultiCam dataset using incomplete 2d- and 3d-pose data. The results show that the proposed OJR method makes our FallNet model robust to missing information in the pose data, and enables the model to make correct fall predictions under 2d or 3d pose errors.
\subsubsection{3d Pose Estimation}
\begin{table}
	\caption{Comparison of the performance of our 3d PoseNet with and without the proposed Occluded Joints Resilience (OJR) on our synthetic dataset under different noise levels.}
	\vspace{-5pt}
	\centering
	\setlength\tabcolsep{10.0pt}\centering
	\begin{tabular}{@{}ccccc@{}}
		\toprule
		\multirow{1}{*}{No. of missing}&\multicolumn{2}{c}{ mean pose error (mm)} \\
		joints&{with OJR}&without OJR\\	
		\midrule
		1&\textbf{17.11}&197.73	\\
		3&\textbf{21.13}&351.19		\\
		5&\textbf{26.25}&420.52		\\
		7&\textbf{34.21}&464.16		\\	
		\bottomrule
	\end{tabular}
	\vspace{0pt}
	\label{table_3d_predictions}
\end{table}
Here, we tested our 3d PoseNet under different levels of noise (missing joints) on the synthetic data. 
For this, we randomly split the data into 70\% train and 30\% test data subsets. Table \ref{table_3d_predictions} shows the mean joints position errors in millimeters which is the mean Euclidean distance between predicted joint positions and ground-truth joint positions averaged over all the joints, produced by our models on the test dataset. 
Table \ref{table_3d_predictions} shows that our OJR-based 3d PoseNet consistently produced lower pose errors compared to the model without OJR for all levels of noise on the test dataset. 
Fig. \ref{fig_human} shows qualitative results of our 3d PoseNet with and without the proposed OJR on our synthetic dataset. The results show that the proposed OJR enables our 3d PoseNet to successfully recover 3d joints information from incomplete 2d pose inputs. This enables our framework to make correct fall predictions under 2d pose errors as shown in Fig. \ref{fig_missing_joints}.
\section{Conclusion and Future Work}
In this paper we present Single Shot Human Fall Detector (SSHFD), a deep learning framework for human fall detection from a single image. SSHFD learns fall representations based on human joint locations in 2d image space and 3d Cartesian space. Our fall representation is invariant to physical appearance, background, and enables our framework to successfully transfer fall recognition knowledge from pure synthetic data to unseen real-world data.
We also present neural network models for 3d pose estimation and fall recognition which are resilient to occluded body parts. Experiments on real-world datasets demonstrate that our framework successfully handles challenging scenes with occlusions. 
These capabilities open new possibilities for advancing human pose based fall detection purely from synthetic data. 
In future, we plan to expand our framework for the recognition of other activities to enhance its potential for general human activity recognition.
\bibliographystyle{ecai}
\bibliography{ijcai19}

\begin{thebibliography}{10}

\bibitem{asifprivacy}
Umar Asif, Benjamin Mashford, Stefan von Cavallar, Shivanthan Yohanandan,
  Subhrajit Roy, Jianbin Tang, and Stefan Harrer, `Privacy preserving human
  fall detection using video data', (2019).

\bibitem{auvinet2010multiple}
Edouard Auvinet, Caroline Rougier, Jean Meunier, Alain St-Arnaud, and
  Jacqueline Rousseau, `Multiple cameras fall dataset', {\em
  DIRO-Universit{\'e} de Montr{\'e}al, Tech. Rep}, {\bf 1350}, (2010).

\bibitem{charfi2013optimized}
Imen Charfi, Johel Miteran, Julien Dubois, Mohamed Atri, and Rached Tourki,
  `Optimized spatio-temporal descriptors for real-time fall detection:
  comparison of support vector machine and adaboost-based classification', {\em
  Journal of Electronic Imaging}, {\bf 22}(4),  041106, (2013).

\bibitem{fleming2008inability}
Jane Fleming and Carol Brayne, `Inability to get up after falling, subsequent
  time on floor, and summoning help: prospective cohort study in people over
  90', {\em Bmj}, {\bf 337},  a2227, (2008).

\bibitem{gasparrini2014depth}
Samuele Gasparrini, Enea Cippitelli, Susanna Spinsante, and Ennio Gambi, `A
  depth-based fall detection system using a kinect sensor', {\em Sensors}, {\bf
  14}(2),  2756--2775, (2014).

\bibitem{Detectron2018}
Ross Girshick, Ilija Radosavovic, Georgia Gkioxari, Piotr Doll\'{a}r, and
  Kaiming He.
\newblock Detectron.
\newblock \url{https://github.com/facebookresearch/detectron}, 2018.

\bibitem{he2016deep}
Kaiming He, Xiangyu Zhang, Shaoqing Ren, and Jian Sun, `Deep residual learning
  for image recognition', in {\em CVPR}, pp. 770--778, (2016).

\bibitem{huang2004extreme}
Guang-Bin Huang, Qin-Yu Zhu, and Chee-Kheong Siew, `Extreme learning machine: a
  new learning scheme of feedforward neural networks', in {\em Neural Networks,
  Proceedings. IEEE International Joint Conference on}, volume~2, pp. 985--990,
  (2004).

\bibitem{hung2013detecting}
Dao~Huu Hung, Hideo Saito, and Gee-Sern Hsu, `Detecting fall incidents of the
  elderly based on human-ground contact areas', in {\em 2nd IAPR Asian
  Conference on Pattern Recognition}, pp. 516--521. IEEE, (2013).

\bibitem{juang2007human}
Chia-Feng Juang and Chia-Ming Chang, `Human body posture classification by a
  neural fuzzy network and home care system application', {\em IEEE
  Transactions on Systems, Man, and Cybernetics-Part A: Systems and Humans},
  {\bf 37}(6),  984--994, (2007).

\bibitem{lin2014microsoft}
Tsung-Yi Lin, Michael Maire, Serge Belongie, James Hays, Pietro Perona, Deva
  Ramanan, Piotr Doll{\'a}r, and C~Lawrence Zitnick, `Microsoft coco: Common
  objects in context', in {\em ECCV}, pp. 740--755, (2014).

\bibitem{martinez2017simple}
Julieta Martinez, Rayat Hossain, Javier Romero, and James~J Little, `A simple
  yet effective baseline for 3d human pose estimation', in {\em Proceedings of
  the IEEE International Conference on Computer Vision}, pp. 2640--2649,
  (2017).

\bibitem{mastorakis2014fall}
Georgios Mastorakis and Dimitrios Makris, `Fall detection system using
  kinect’s infrared sensor', {\em Journal of Real-Time Image Processing},
  {\bf 9}(4),  635--646, (2014).

\bibitem{miaou2006customized}
S-G Miaou, Pei-Hsu Sung, and Chia-Yuan Huang, `A customized human fall
  detection system using omni-camera images and personal information', in {\em
  Distributed Diagnosis and Home Healthcare, 1st Transdisciplinary Conference
  on}, pp. 39--42. IEEE, (2006).

\bibitem{mirmahboub2013automatic}
Behzad Mirmahboub, Shadrokh Samavi, Nader Karimi, and Shahram Shirani,
  `Automatic monocular system for human fall detection based on variations in
  silhouette area', {\em IEEE Transactions on Biomedical Engineering}, {\bf
  60}(2),  427--436, (2013).

\bibitem{newell2016stacked}
Alejandro Newell, Kaiyu Yang, and Jia Deng, `Stacked hourglass networks for
  human pose estimation', in {\em ECCV}, pp. 483--499, (2016).

\bibitem{paszke2017automatic}
Adam Paszke, Sam Gross, Soumith Chintala, Gregory Chanan, Edward Yang, Zachary
  DeVito, Zeming Lin, Alban Desmaison, Luca Antiga, and Adam Lerer, `Automatic
  differentiation in pytorch', (2017).

\bibitem{qian2008home}
Huimin Qian, Yaobin Mao, Wenbo Xiang, and Zhiquan Wang, `Home environment fall
  detection system based on a cascaded multi-svm classifier', in {\em ICARCV},
  pp. 1567--1572. IEEE, (2008).

\bibitem{zhao2012exploration}
Guoru Zhao, Zhanyong Mei, Ding Liang, Kamen Ivanov, Yanwei Guo, Yongfeng Wang,
  and Lei Wang, `Exploration and implementation of a pre-impact fall
  recognition method based on an inertial body sensor network', {\em Sensors},
  {\bf 12}(11),  15338--15355, (2012).

\end{thebibliography}
%
\end{document}